\documentclass{article}

\PassOptionsToPackage{table}{xcolor}
\usepackage{etoolbox}       %
\usepackage[utf8]{inputenc} %

\usepackage{graphicx}  %

\usepackage{amsmath,amsfonts,bm}

\def\eqref#1{equation~(\ref{#1})}

\def\1{\bm{1}}

\DeclareMathAlphabet{\mathsfit}{\encodingdefault}{\sfdefault}{m}{sl}
\SetMathAlphabet{\mathsfit}{bold}{\encodingdefault}{\sfdefault}{bx}{n}

\newcommand{\mahdi}[1]{}

\def\remark{\addtocounter{remark}{1}\def\@currentlabel{\theremark}%
\emph{Remark~\theremark}. } \makeatother
\newcounter{remark}

 \usepackage{color-edits}
 \addauthor{mh}{red}
 \addauthor{xw}{magenta}
 \addauthor{ne}{brown}
 
\usepackage{anyfontsize}

\usepackage[parfill]{parskip}
\usepackage{amsmath,amsthm}
\usepackage{mathtools}  %
\usepackage{adjustbox}
\usepackage{threeparttable}
\usepackage{tablefootnote}
\usepackage{afterpage}

  \usepackage[parfill]{parskip}
  \usepackage[sort&compress,numbers]{natbib}

  \newlength{\defbaselineskip}
  \RequirePackage[T1]{fontenc}
  \RequirePackage[tt=false, type1=true]{libertine}
  \RequirePackage[varqu]{zi4}
  \RequirePackage[libertine]{newtxmath}

\usepackage{bm}

\usepackage[inline]{enumitem}
\usepackage{booktabs}
\usepackage{subcaption}
\usepackage{multirow}
\usepackage{wrapfig}
\usepackage{algorithm,algorithmicx,algpseudocode}
\usepackage{nicematrix}

\usepackage{bbm}
\usepackage{float}
\usepackage{multicol}

\usepackage{import}
\usepackage{lettrine}

\usepackage{pifont}%

\usepackage[T1]{fontenc}

\usepackage{thm-restate}

\usepackage{url}            %
\usepackage{booktabs}       %
\usepackage{amsfonts}       %
\usepackage{nicefrac}       %
\usepackage{microtype}      %

\usepackage{wrapfig}

\usepackage{caption}
\usepackage{capt-of}
\usepackage{lipsum}

\usepackage[colorlinks]{hyperref}
\usepackage[capitalise,noabbrev]{cleveref}

\usepackage{threeparttable}

\definecolor{c1}{HTML}{586770}
\definecolor{c4}{HTML}{2a4a67}
\definecolor{c3}{HTML}{6d2a58}
\definecolor{c2}{HTML}{34142a}
\hypersetup{colorlinks=true, citecolor=c3, linkcolor=c2, urlcolor=c2}

\definecolor{myblue}{HTML}{FDF5E0}
\definecolor{mygray}{HTML}{DBE2E9}
\definecolor{mygreen}{HTML}{E6F3FC}

\definecolor{dark2orange}{rgb}{0.9, 0.4, 0.}
\definecolor{dark2purple}{rgb}{0.4, 0.4, 0.8}

\hypersetup{
    colorlinks=true,
    linkcolor=black,
    urlcolor=c1,
    citecolor=c1,
}

\usepackage{multirow}

\usepackage{mathtools}

\usepackage{tikz}

\usepackage{authblk}
\usepackage{epigraph}
\usepackage{soul}

\usepackage{tcolorbox}
\usepackage{xcolor}
\newtcolorbox{c4box}{boxrule=1pt, colback=c4!5!white,colframe=c4!50!white}
\newtcolorbox{c3box}{boxrule=1pt, colback=c3!5!white,colframe=c3!50!white}
\newtcolorbox{c2box}{boxrule=1pt, colback=c2!5!white,colframe=c2!50!white}
\newtcolorbox{c1box}{boxrule=1pt, colback=c1!5!white,colframe=c1!50!white}

\newcommand\blfootnote[1]{%
  \begingroup
  \renewcommand\thefootnote{}\footnote{#1}%
  \addtocounter{footnote}{-1}%
  \endgroup
}

\title{Grow, Don't Overwrite: Fine-tuning Without Forgetting}

\author{%
Dyah Adila$^{1}$\thanks{Work done during internship at Google Research.} \quad Hanna Mazzawi$^{2}$ \quad Benoit Dherin$^{2}$ \quad Xavier Gonzalvo$^{2}$ \\
$^1$University of Wisconsin-Madison \quad $^2$Google Research
\protect\blfootnote{Corresponding author: Dyah Adila, \texttt{adila@wisc.edu}}
}

\date{}

\begin{document}

\vspace{-20ex}
\maketitle

\begin{abstract}
  Adapting pre-trained models to specialized tasks often leads to catastrophic forgetting, where new knowledge overwrites foundational capabilities. Existing methods either compromise performance on the new task or struggle to balance training stability with efficient reuse of pre-trained knowledge. We introduce a novel function-preserving expansion method that resolves this dilemma. Our technique expands model capacity by replicating pre-trained parameters within transformer submodules and applying a scaling correction that guarantees the expanded model is mathematically identical to the original at initialization, enabling stable training while exploiting existing knowledge. Empirically, our method eliminates the trade-off between plasticity and stability, matching the performance of full fine-tuning on downstream tasks without any degradation of the model's original capabilities. Furthermore, we demonstrate the modularity of our approach, showing that by selectively expanding a small subset of layers we can achieve the same performance as full fine-tuning at a fraction of the computational cost.
\end{abstract}

 \section{Introduction}

\label{sec:intro}

Neural networks suffer from a severe degradation of previously learned knowledge when fine-tuned on a new task, a phenomenon known as \textbf{\textit{catastrophic forgetting}} \citep{mccloskey1989catastrophic, kirkpatrick2017overcoming, ramasesh2020anatomy, de2021continual}. This poses a critical obstacle for adapting large pre-trained models to new, specialized domains. When a model is fine-tuned for a specific purpose, such as scientific reasoning or medical diagnosis, it risks losing the foundational knowledge that makes it powerful. For example, a model specialized in quantum physics literature may excel at generating novel hypotheses but lose its ability to perform basic arithmetic, fundamentally limiting its utility and forcing practitioners into a compromise between expert skill and general competence \cite{legg2007universal, dong2023abilities}.

 

This forgetting happens because standard optimization overwrites a model's parameters to fit new data, causing its internal representations to shift abruptly and erase existing skills \cite{ramasesh2020anatomy, jiang2025unlocking}. A common family of solutions uses regularization, which adds a penalty to the loss function to discourage the model from deviating too far from its original state \citep{kirkpatrick2017overcoming, ramasesh2020anatomy, jiang2025unlocking}. 

However, this imposes a zero-sum trade-off: within a fixed-capacity model, any resource allocated to remembering the past is a resource taken away from learning the future, preventing the model from excelling at either \cite{grossberg2012studies}.

\begin{figure}[htp!]
        \centering
        \begin{subfigure}[b]{0.49\textwidth}
            \centering
            \includegraphics[width=\textwidth]{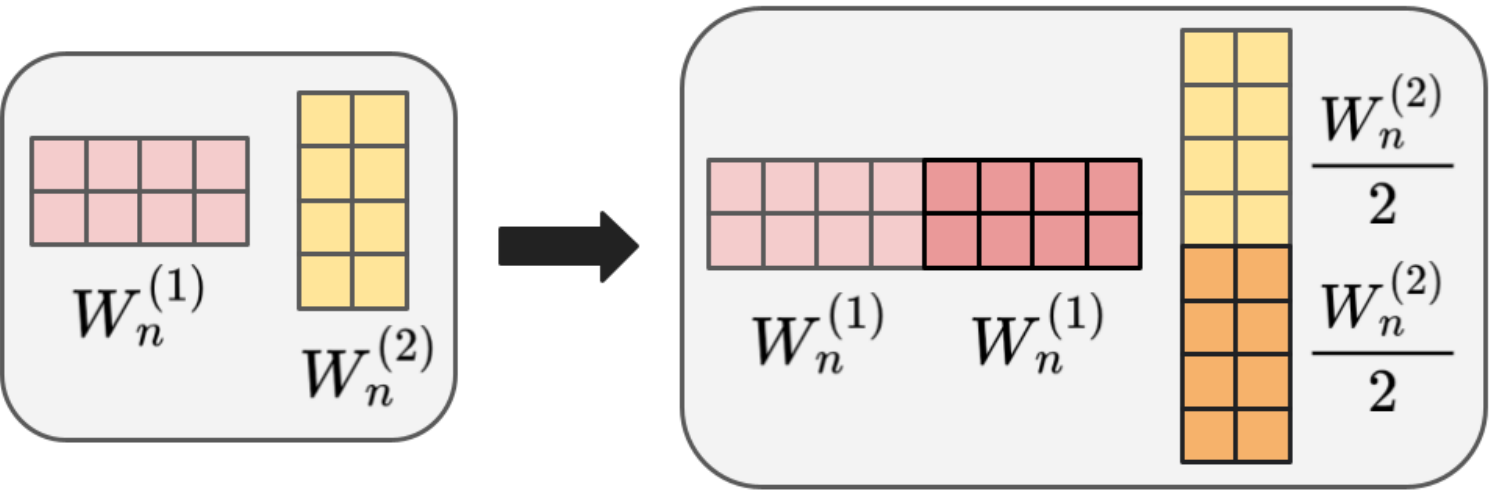}
            \caption[]%
            {{\small Our growing approach}}    
            \label{fig:approach}
        \end{subfigure}
        \hfill
        \begin{subfigure}[b]{0.49\textwidth}   
            \centering 
            \includegraphics[width=\textwidth]{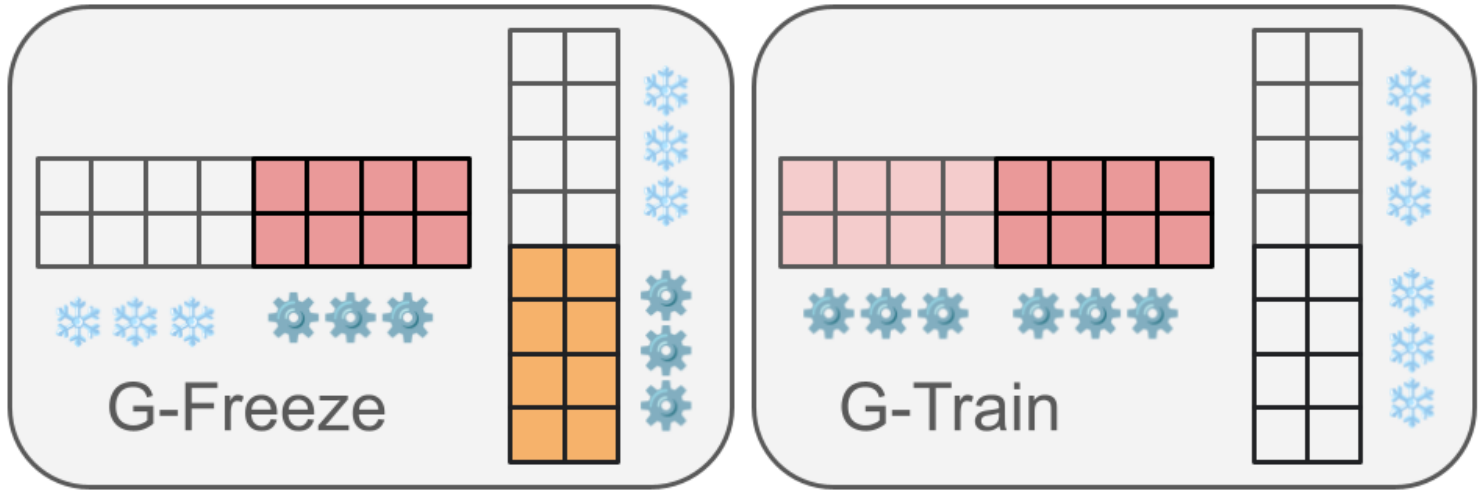}
            \caption[]%
            {{\small Fine-tuning Strategy}}    
            \label{fig:variants}
        \end{subfigure}
        \caption[]
        {\small \textbf{(a)} We double the MLP's hidden dimension by duplicating the up-projection weights ($W_n^{(1)}$) and compensating in the down-projection layer ($W_n^{(2)}$) to preserve the original function. \textbf{(b)} In the \textbf{G-Freeze} variant, only new parameters (darker shades) are trained. In the \textbf{G-Train} variant, the entire up-projection matrix is trained while the down-projection matrix is frozen, as indicated by the snowflake symbol.} 
        \label{fig:main_figure}
    \end{figure}

An alternative family of methods, capacity growth, circumvents this trade-off by adding new parameters for new tasks while freezing the original model. This, however, introduces its own dilemma between stability and efficiency. For training to be stable, the expansion must be function-preserving--that is, it must not alter the model's outputs on existing data upon initialization \citep{gesmundo2023composable}. Yet, for learning to be efficient, new components should exploit the rich knowledge from pre-training rather than starting from scratch. Current methods fail to reconcile these two needs. Some achieve stability by inserting randomly initialized identity modules, which is inefficient as it ignores existing knowledge \citep{houlsby2019parameter, wu2024llama, han2025loire}. Others attempt to reuse pre-trained weights but violates the function-preserving constraint \citep{wei2025control}. A method that can expand capacity using pre-trained knowledge while guaranteeing function-preservation has remained a critical gap in the literature.

In this work, we close this gap with a function-preserving expansion technique inspired by Deep Fusion \citep{mazzawi2023deep}. The core of our method is to add capacity by replicating entire computational units within a Transformer's MLP submodules (i.e., the intermediate neurons in the MLP). We guarantee that our approach is function-preserving with a two-step process occurring within each submodule. First, we expand its internal dimension by creating $k$ copies of the parameters in its up-projection layer (which maps the input to a wider hidden state). Second, to ensure the final output remains unchanged, we apply a compensatory scaling to the weights of the down-projection layer (which maps the internal state back to the model's hidden dimension), dividing them by the replication factor $k$. This simple scaling perfectly counteracts the wider internal activation, ensuring the expanded model is mathematically identical to the original at initialization.

Our experiments show this approach resolves the trade-off between learning new skills and retaining old ones. Across a diverse set of benchmarks---including question-answering, translation, and mathematical reasoning---our method \textbf{matches the performance of standard fine-tuning} on new tasks while often exhibiting \textbf{almost zero performance degradation} on the model's original capabilities. We also demonstrate that expanding only a targeted subset of layers is sufficient to match the results of a network-wide expansion, substantially reducing computational cost. Furthermore, because it only expand MLP submodules, even a full expansion of every layer requires training \textbf{only  60\% of the original model's parameters}, offering a considerable saving compared to standard fine-tuning, which modifies 100\%.

\paragraph{Main Contributions:}

\begin{itemize}[leftmargin=*]

\item A novel function-preserving network growing method that reuses pre-trained knowledge to learn new skills.

\item Our approach matches standard fine-tuning (SFT) performance on new tasks while completely eliminating catastrophic forgetting.

\item A modular framework where this performance is achieved by expanding and finetuning only a targeted subset of layers, reducing computational cost.

\end{itemize}


\section{Background}
\label{sec:background}
\paragraph{Problem Statement.}
Given a pretrained language model $M_0$ with parameters $\theta_0$, we address the task of fine-tuning it for a downstream task represented by the dataset $\mathcal{D}_{T}$. The objective is to produce an adapted model, $M_T$ with parameters $\theta_T$, that minimizes the new task loss without degrading performance on the original pre-training distribution. Our goal can be formally expressed as:
\[
\min_{\theta_T} \mathcal{L}_T(\theta_T) \quad \text{subject to} \quad \mathcal{L}_{PT}(\theta_T) \le \mathcal{L}_{PT}(\theta_0)
\]
In practice, the full pre-training corpus $\mathcal{D}_{PT}$ is often inaccessible. We therefore approximate the model's general capabilities by measuring its performance on a proxy benchmark, $\mathcal{D}_{\text{proxy}}$, which covers a broad range of foundational skills. Our evaluation thus aims to confirm low test error on the new task $\mathcal{D}_{T}$ while demonstrating no increase in test error on the proxy benchmark $\mathcal{D}_{\text{proxy}}$ relative to the original model $M_0$.

\paragraph{Preliminaries.}
We first establish notation for the Transformer architecture, adopting a convention similar to \cite{gesmundo2023composable}. A Transformer model is composed of a stack of $N$ layers. Each layer $n \in [1, N]$ contains two primary submodules: a multi-head self-attention (MHA) mechanism and multi-layer perceptron (MLP), with residual connections and layer normalization applied after each.

Our approach grows the MLP submodules within Transformer models. The operation of a standard MLP in layer $n$ is defined as:
\begin{align*}
\text{MLP}_n(\mathbf{X}) &= \text{ReLU}(\mathbf{X}\mathbf{W}_n^{(1)} + \mathbf{B}_n^{(1)}) \times \mathbf{W}_n^{(2)} + \mathbf{B}_n^{(2)}
\end{align*}
The input to the MLP submodule is $\mathbf{X} \in \mathbb{R}^{s \times h}$, where $s$ is the sequence length and $h$ is the model's hidden dimension. The MLP consists of two linear layers. The first is an up-projection layer ($\mathbf{W}_n^{(1)} \in \mathbb{R}^{h \times p}$ and $\mathbf{B}_n^{(1)}$) that maps the input from dimension $h$ to a usually larger intermediate hidden dimension $p$. The second is a down-projection layer ($\mathbf{W}_n^{(2)} \in \mathbb{R}^{p \times h}$ and $\mathbf{B}_n^{(2)}$) that projects the output back to dimension $h$ for the subsequent Transformer layer.

For brevity in subsequent sections, we omit the bias terms ($\mathbf{B}_n^{(1)}, \mathbf{B}_n^{(2)}$) and use underset notation to denote matrix dimensions (i.e., writing $\mathbf{W}_n^{(1)} \in \mathbb{R}^{h \times p}$ as $\underset{h \times p}{\mathbf{W}_n^{(1)}}$). Our method, however, remains fully compatible with models that include bias terms.

\section{Methodology}
\label{sec:method}
Our method adds capacity to a pre-trained Transformer by growing its MLP submodules in a function-preserving manner. This guarantees the model's initial behavior remains unchanged, while allowing it to learn new skills by building upon pretrained parameters instead of starting from random initialization. We achieve this with a two-step process.

First, we double the MLP's  hidden dimension $p$ by duplicating the up-projection weight matrix, $\mathbf{W}_n^{(1)}$. The new expanded matrix, $\hat{\mathbf{W}}_n^{(1)}$, is formed by horizontally concatenating the original matrix with itself:
\[
\underset{h \times p}{\mathbf{W}_n^{(1)}} \mapsto \underset{h \times 2p}{\hat{\mathbf{W}}_{n}^{(1)}} := \left[ \underset{h \times p}{\mathbf{W}_n^{(1)}} \quad \underset{h \times p}{\mathbf{W}_n^{(1)}}\right]
\]
Next, to ensure the output of the MLP submodule remains identical, we compensate for this expansion in the down-projection layer. The new down-projection matrix, $\hat{\mathbf{W}}_n^{(2)}$, is formed by vertically concatenating the original matrix, scaled by a factor of $\frac{1}{2}$, with itself:
\[
\underset{p \times h}{\mathbf{W}_n^{(2)}} \mapsto \underset{2p \times h}{\hat{\mathbf{W}}_n^{(2)}} := \begin{bmatrix} \frac{1}{2}\underset{p \times h}{\mathbf{W}_n^{(2)}} \\ \frac{1}{2}\underset{p \times h}{\mathbf{W}_n^{(2)}} \end{bmatrix}
\]

\begin{proof}[Proof of Function Preservation]
Let $\mathbf{Y} = \text{ReLU}(\mathbf{X}\mathbf{W}_n^{(1)})$. The output of the original, non-grown MLP is $\mathbf{Y}\mathbf{W}_n^{(2)}$. After expansion, the output of the up-projection layer is $[\mathbf{Y} \quad \mathbf{Y}]$. The final output of the expanded MLP is thus:
\[
[\mathbf{Y} \quad \mathbf{Y}] \begin{bmatrix} \frac{1}{2}\mathbf{W}_n^{(2)} \\ \frac{1}{2}\mathbf{W}_n^{(2)} \end{bmatrix} = \frac{1}{2}\mathbf{Y}\mathbf{W}_n^{(2)} + \frac{1}{2}\mathbf{Y}\mathbf{W}_n^{(2)} = \mathbf{Y}\mathbf{W}_n^{(2)}
\]
This is identical to the output of the original MLP.
\end{proof}

\paragraph{Fine-tuning Strategy.}
\textbf{We freeze all original model parameters} and exclusively train the new ones, enabling the model to learn new skills without disrupting its original capabilities. While standard fine-tuning of a 1B-parameter model updates all weights, our method—even when growing all layers—trains only the new parameters, which total approximately \textasciitilde{}60\% of the original count. As we show in Section~\ref{sec:exp_layer_subset}, this cost can be reduced further by growing only a targeted subset of layers, often with comparable performance.

From this principle, we derive two fine-tuning variants, detailed in Figure~\ref{fig:variants}:
\begin{itemize}[leftmargin=*]
    \item \textbf{G-Freeze}: Our primary and default strategy. It strictly trains \textbf{only the newly added weights} in both the up-projection ($\hat{\mathbf{W}}_n^{(1)}$) and down-projection ($\hat{\mathbf{W}}_n^{(2)}$) matrices.
    \item \textbf{G-Train}: An alternative strategy designed for cognitively demanding tasks like mathematical reasoning. This variant involves fine-tuning the \textbf{entire expanded up-projection matrix} ($\hat{\mathbf{W}}_n^{(1)}$) while keeping the \textbf{entire down-projection matrix} ($\hat{\mathbf{W}}_n^{(2)}$) and all other original parameters frozen. This approach is informed by prior work suggesting that factual knowledge is localized in the down-projection layer, which we aim to preserve \citep{geva2020transformer, meng2022locating}.
\end{itemize}
Unless otherwise stated, all experiments use the \textbf{G-Freeze} strategy.

\paragraph{Generalization to Arbitrary Expansion Factors.}
Our approach can be generalized from two copies to an arbitrary integer factor, $k \ge 2$. In this case, the up-projection matrix is concatenated $k$ times, and each of the $k$ copies in the stacked down-projection matrix is scaled by $\frac{1}{k}$. The function-preserving property holds for any value of $k$. While our method supports any expansion factor, we found empirically that using $k=2$ consistently provided the best trade-off between performance gains and parameter efficiency. We therefore use $k=2$ for all experiments in this paper.

\begin{figure}[htp!]
        \centering
        \begin{subfigure}[b]{0.49\textwidth}
            \centering
            \includegraphics[width=\textwidth]{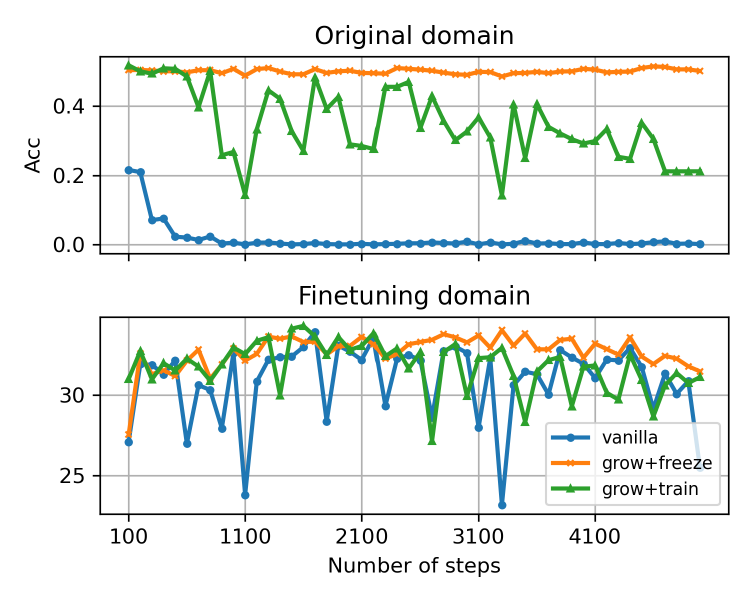}
            \caption[]%
            {{\small French translation}}    
            \label{fig:main_mtnt}
        \end{subfigure}
        \begin{subfigure}[b]{0.49\textwidth}  
            \centering 
            \includegraphics[width=\textwidth]{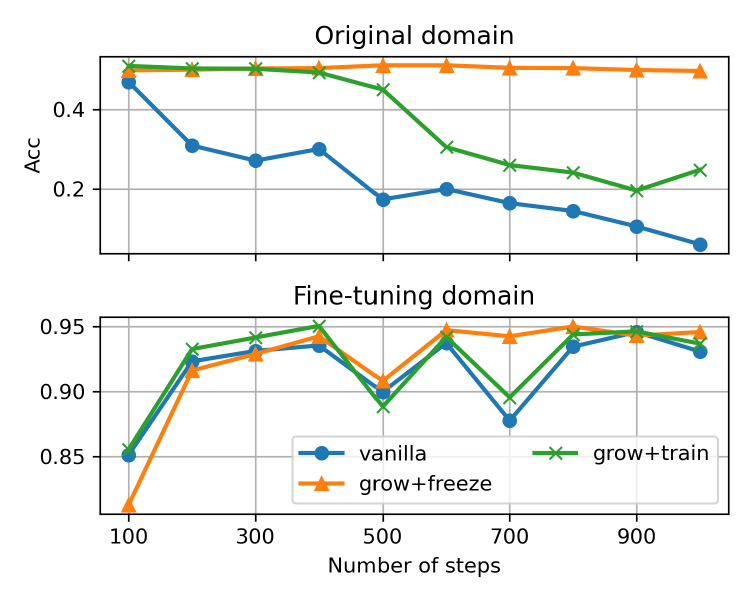}
            \caption[]%
            {{\small Science entailment}}    
            \label{fig:main_scitail}
        \end{subfigure}
        \begin{subfigure}[b]{0.49\textwidth}   
            \centering 
            \includegraphics[width=\textwidth]{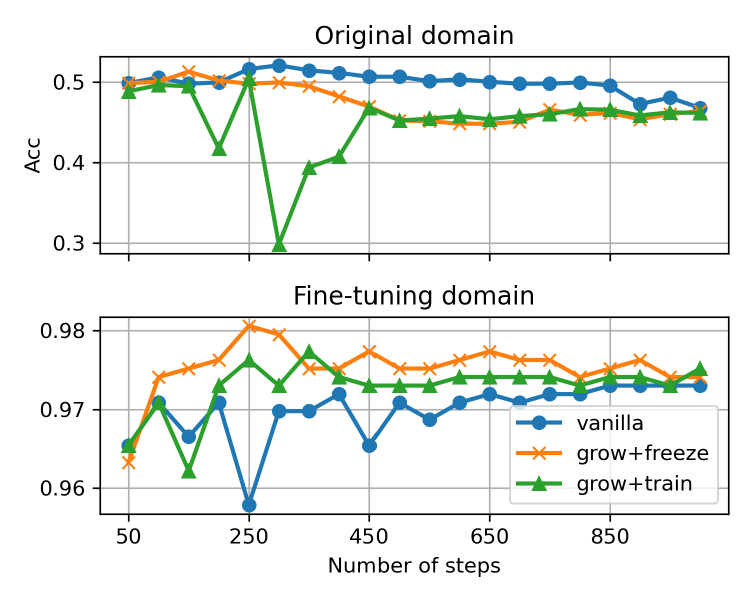}
            \caption[]%
            {{\small Science Q\&A}}    
            \label{fig:main_qasc}
        \end{subfigure}
        \begin{subfigure}[b]{0.49\textwidth}   
            \centering 
            \includegraphics[width=\textwidth]{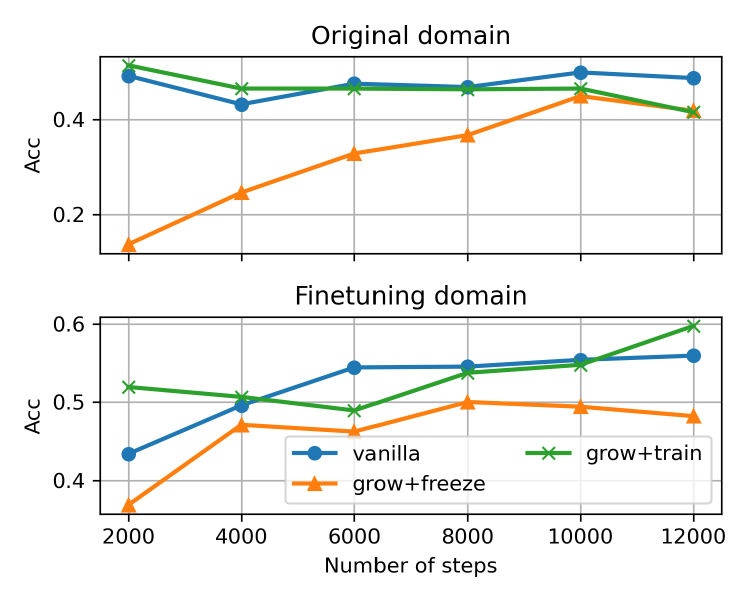}
            \caption[]%
            {{\small Math QA}}    
            \label{fig:main_math}
        \end{subfigure}
        \caption[]
        {\small SFT (blue) shows severe degradation on the original domain (top plots), particularly for tasks with large domain shifts like translation and entailment. Our method (green, orange) maintain original performance while matching or exceeding the baseline on the new fine-tuning tasks (bottom plots).}
        \label{fig:result_1b}
    \end{figure}
\begin{figure}[ht!]
        \centering
        \begin{subfigure}[b]{0.32\textwidth}
            \centering
            \includegraphics[width=\textwidth]{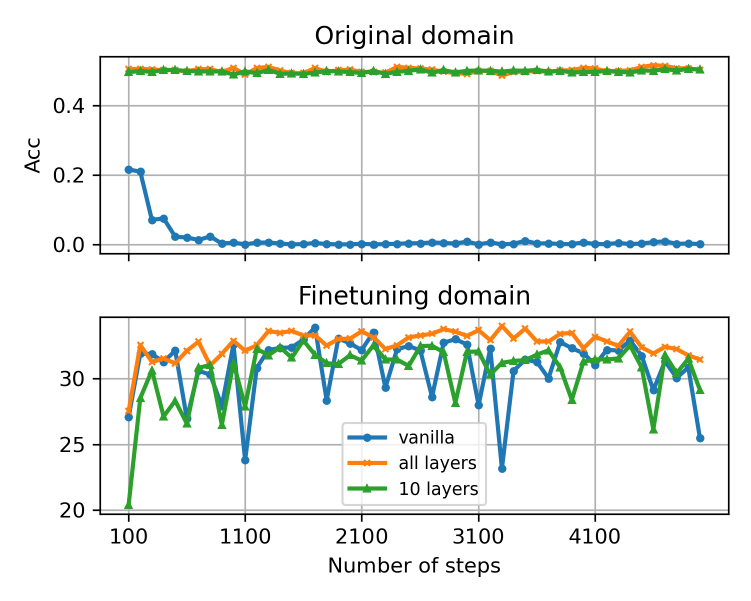}
            \caption[]%
            {{\small French translation}}    
            \label{fig:layer_subset_mtnt}
        \end{subfigure}
        \begin{subfigure}[b]{0.32\textwidth}  
            \centering 
            \includegraphics[width=\textwidth]{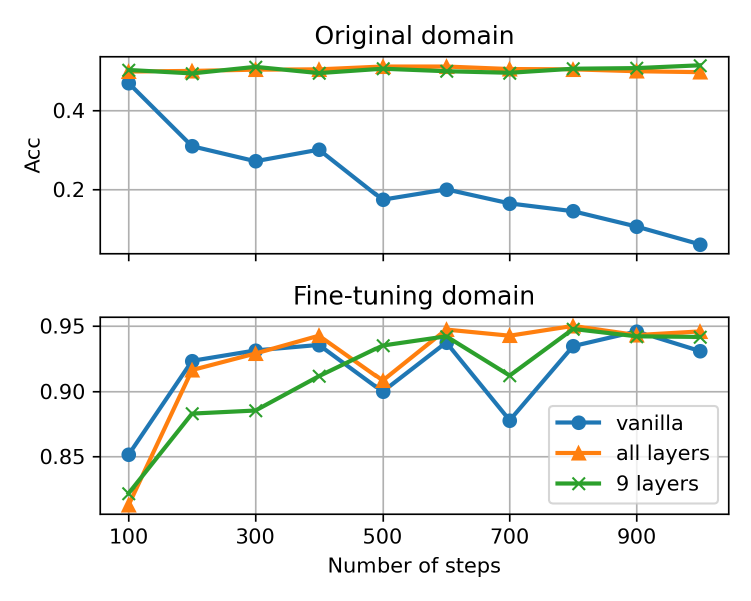}
            \caption[]%
            {{\small Science entailment}}    
            \label{fig:layer_subset_scitail}
        \end{subfigure}
        \begin{subfigure}[b]{0.32\textwidth}   
            \centering 
            \includegraphics[width=\textwidth]{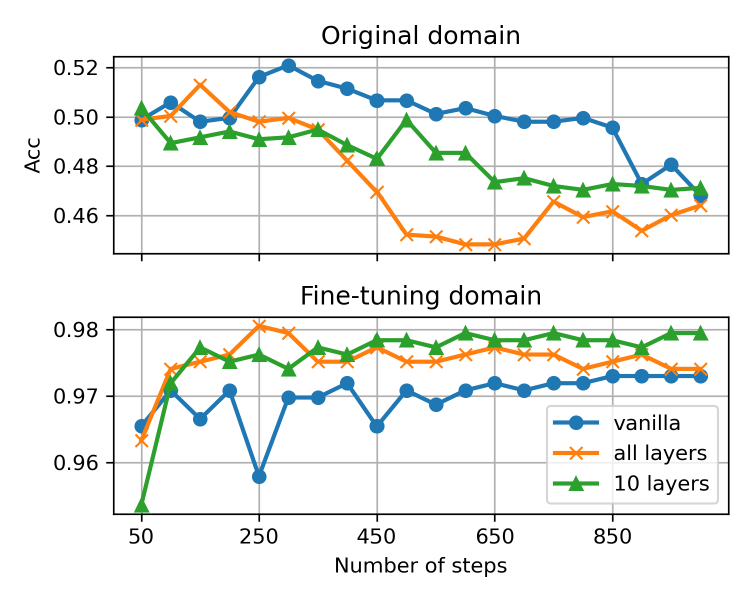}
            \caption[]%
            {{\small Science Q\&A}}    
            \label{fig:layer_subset_qasc}
        \end{subfigure}
        \caption[]
        {\small Full performance can be achieved with a fraction of the trainable parameters. Growing a targeted subset of 10 layers (green) consistently matches the performance of growing all layers (orange).} 
        \label{fig:layer_subset}
    \end{figure}

\section{Experimental Results}
\label{sec:experiments}
This section evaluates the following claims:

\begin{itemize}[leftmargin=*]






\item \textbf{Resolving the Forgetting-Performance Trade-off (Section \ref{sec:exp_remove_forgetting}).} Our method matches full SFT performance on new tasks while completely eliminating catastrophic forgetting.

\item \textbf{Parameter Efficiency (Section \ref{sec:exp_layer_subset}).} Expanding a small, targeted subset of layers achieves the performance of growing the entire model at a fraction of the computational cost.

\item \textbf{Scalable Improvement (Section \ref{sec:exp_scaling}).} Performance on new tasks scales positively with the number of layers expanded, linking increased capacity to improved skill.

\item \textbf{Task Complexity (Section \ref{sec:exp_rank}).} Capacity correlates with task difficulty; more cognitively demanding tasks like mathematical reasoning require larger expansions for optimal performance.

\item \textbf{Representation Stability (Section \ref{sec:exp_fv}).} Our method preserves the model's original latent representations, directly preventing the representational shift known to cause forgetting.

\end{itemize}

\paragraph{Datasets and Metrics.}
We measure performance on two fronts: knowledge retention and new task acquisition. To measure retention and quantify catastrophic forgetting, we follow recent work \citep{wei2025control, jiang2025unlocking} by evaluating performance on commonsense reasoning and language understanding benchmarks--a task that reflects foundational knowledge acquired during pre-training. We use the WinoGrande schema challenge \citep{sakaguchi2019winogrande}.

To measure new task acquisition, we test on four tasks with varying degrees of domain shift from the general pre-training distribution: tasks with significant shifts, such as English-French translation (mtnt; \citealp{michel2018mtnt}) and entailment with its rigid answer format (SciTail; \citealp{khot2018scitail}), and tasks with more subtle shifts, like science question-answering (QASC; \citealp{allenai:qasc}) and mathematical reasoning (MathQA; \citealp{amini2019mathqa}). Performance on mtnt is measured using SacreBLEU \citep{post-2018-call}, while for all other tasks, we report accuracy based on an exact string match between the model's output and the ground truth answer. Unless stated otherwise, experiments on MathQA are using our G-train variant, and all other datasets are using G-freeze.

\paragraph{Setup.} For all experiments, we finetune Gemma3-1B \citep{team2025gemma} models using Adam optimizer with $1e-3$ learning rate. To ensure a fair comparison, we determine the number of training steps based on the convergence of the vanilla finetuning baseline for each respective task. We then train all models for this same number of updates, plotting performance on both the new task and the WinoGrande proxy benchmark. This allows us to visualize the dynamics of new skill acquisition versus knowledge retention over time.

\subsection{Eliminates Forgetting}
\label{sec:exp_remove_forgetting}

\paragraph{Results.}
Figure \ref{fig:result_1b} shows that \textbf{our approach successfully eliminates the trade-off} between new task acquisition and catastrophic forgetting. Our G-Freeze variant (orange line) achieves downstream performance comparable or superior to standard fine-tuning while almost perfectly preserving original domain knowledge. In stark contrast, the baseline (blue line) suffers a complete collapse of prior knowledge, with its accuracy on the original domain plummeting to near-zero on tasks with large domain shifts (Figures \ref{fig:main_mtnt} and \ref{fig:main_scitail}). Our results hold for larger models, as shown in Appendix \ref{sec:appendix_extra_exp}.

On more cognitively demanding tasks like MathQA (Figure \ref{fig:main_math}), the G-Train variant (green line)—which fine-tunes all up-projection parameters—outperforms G-Freeze. We hypothesize that for complex tasks with only a moderate domain shift, unfreezing the original up-projection parameters ($\textbf{W}^{(1)}$) provides the necessary plasticity to boost new task performance without risking the loss of foundational knowledge stored in the down-projection layer. In Appendix \ref{sec:exp_scale}, we show that our results persist in bigger models.

\subsection{Matching Full Fine-tuning with a Fraction of the Parameters}
\label{sec:exp_layer_subset}
\paragraph{Setup.} To evaluate the modularity and efficiency of approach, we compare the performance of growing all layers versus growing only a targeted subset. We identify the most task-relevant layers using a simple heuristic: after a preliminary standard fine-tuning run, we rank layers by the magnitude of their weight updates and select the top-$N$ for expansion. While we use this straightforward method for our analysis, we hypothesize that performance could be further optimized with more sophisticated skill localization techniques \citet{panigrahi2023task, todd2024function}.

\paragraph{Results.} Figure \ref{fig:layer_subset} shows that \textbf{growing a targeted subset of 9-10 layers consistently matches the performance of growing all layers} (green vs. orange lines, respectively). This allows us to achieve full performance while nearly halving the number of trainable parameters, reducing them from \textasciitilde{}60\% of the full model to \textasciitilde{}30\%.

\begin{figure}[htp!]
    \begin{subfigure}{0.49\textwidth}
        \centering
        \includegraphics[width=\textwidth]{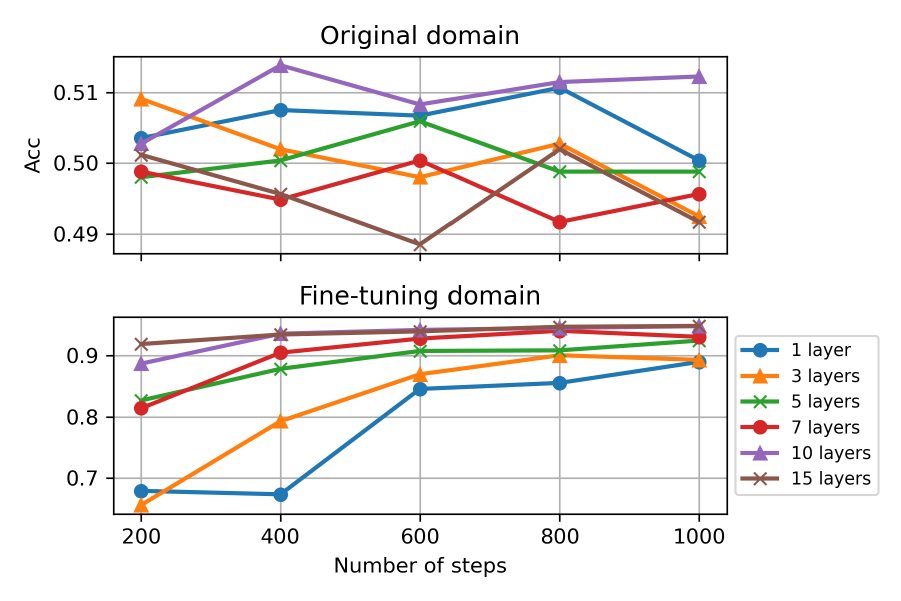}
        \caption[]%
        {{\small G-freeze: Science entailment}}    
        \label{fig:ablation_scitail}
    \end{subfigure}
    \begin{subfigure}{0.49\textwidth}  
        \centering 
        \includegraphics[width=\textwidth]{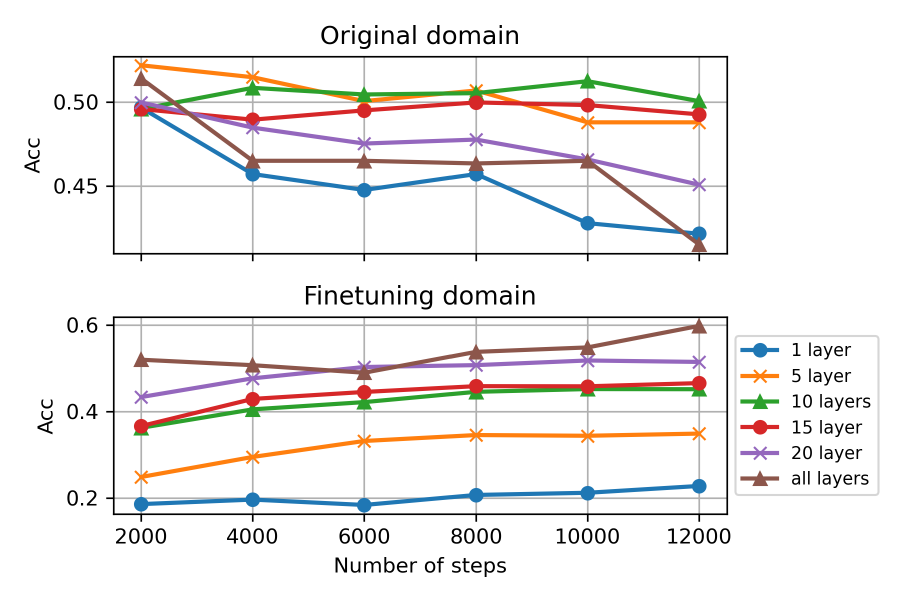}
        \caption[]%
        {{\small G-train: Math QA}}    
        \label{fig:ablation_math}
    \end{subfigure}
    \caption[]
    {\small Performance scales with the number of grown layers $N$. New task performance (bottom) improves as $N$ increases, an effect most significant on the more complex MathQA task (b).} 
    \label{fig:ablations}
\end{figure}

\subsection{Scaling Properties}
\label{sec:exp_scaling}
\paragraph{Setup.} Next, we use the same layer selection procedure as Section \ref{sec:exp_layer_subset}, and vary $N \in \{ 1, 3, 5, 7, 10, 15, 20 \}$ to test how performance changes with growing more layers.
\paragraph{Results.} Figure~\ref{fig:ablations} shows that \textbf{new task performance scales positively with the number of layers grown ($N$)}. This effect is most pronounced on complex tasks like MathQA (Figure~\ref{fig:ablation_math}), whereas on simpler tasks like entailment (Figure~\ref{fig:ablation_scitail}), the benefit is concentrated in the early training stages before performance converges. This scaling introduces a minor forgetting trade-off for the G-Train variant on MathQA, where increasing $N$ causes a gradual decline in original domain performance. The G-Freeze variant, however, does not exhibit this trade-off, with its original domain accuracy remaining stable within a $\pm2\%$ range.

\begin{figure}[ht!]
    \begin{subfigure}{0.32\textwidth}
        \centering
        \includegraphics[width=\textwidth]{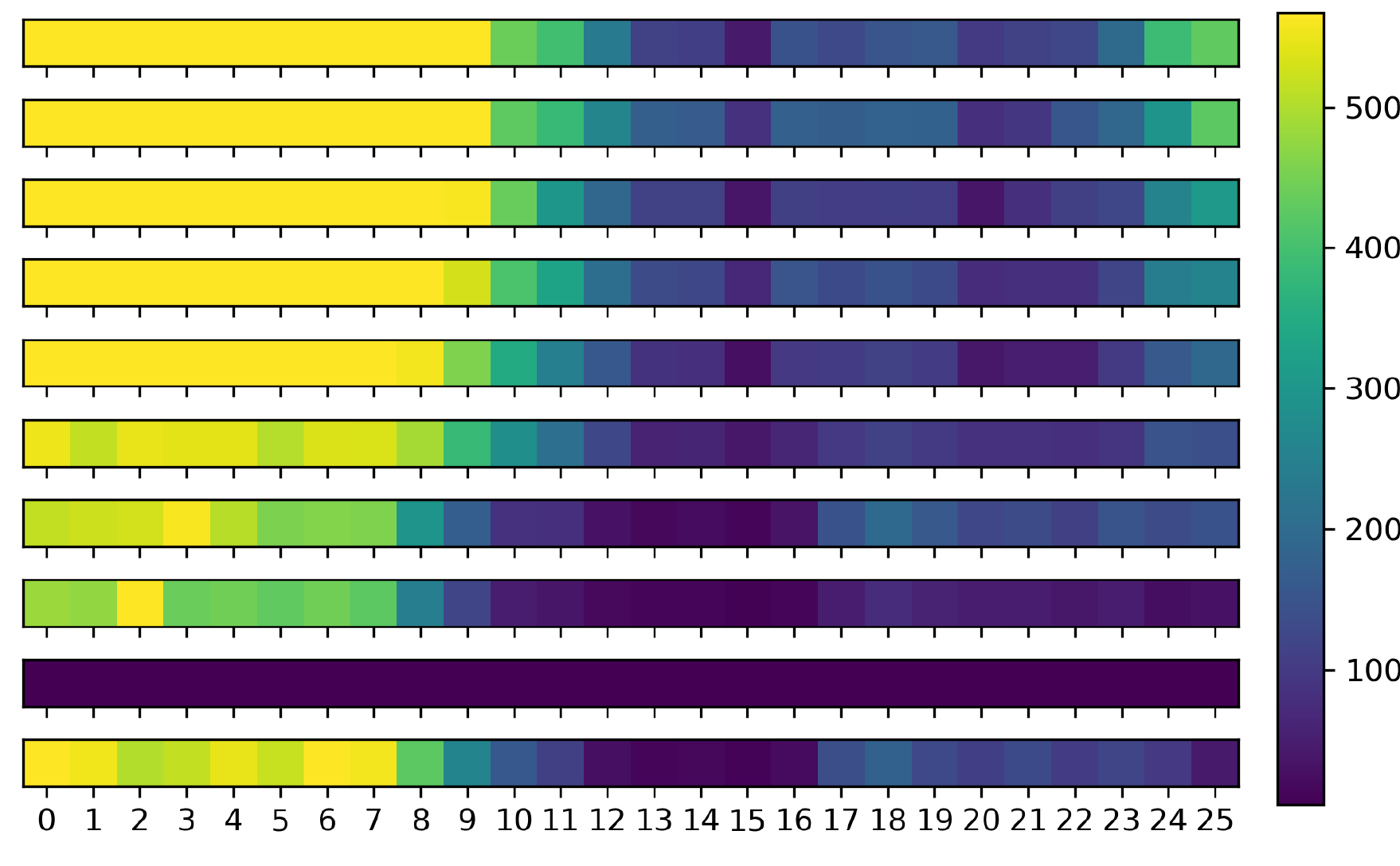}
        \caption[]%
        {{\small Science entailment}}    
        \label{fig:mean and std of net14}
    \end{subfigure}
    \begin{subfigure}{0.32\textwidth}  
        \centering 
        \includegraphics[width=\textwidth]{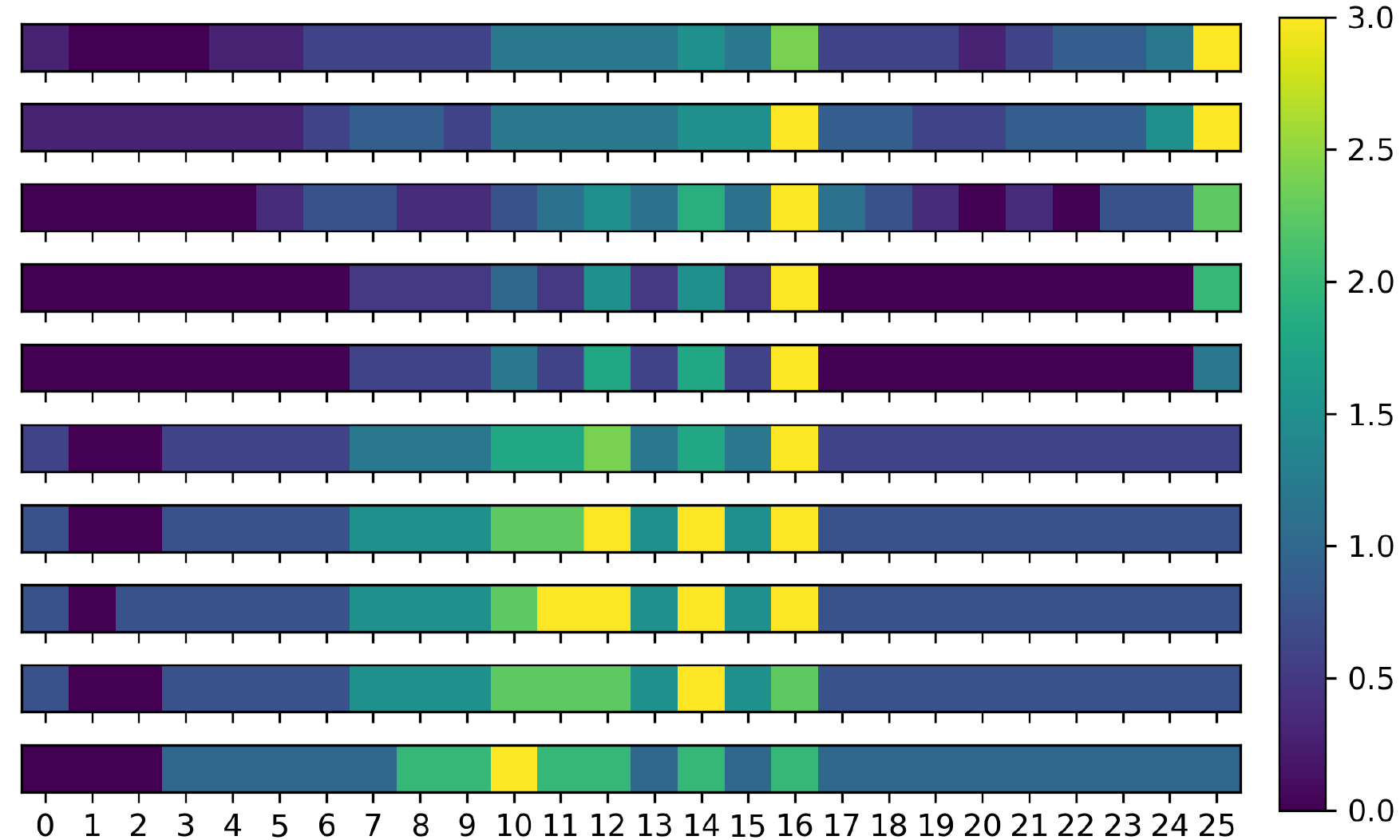}
        \caption[]%
        {{\small French Translation}}    
        \label{fig:mean and std of net24}
    \end{subfigure}
    \begin{subfigure}{0.32\textwidth}   
        \centering 
        \includegraphics[width=\textwidth]{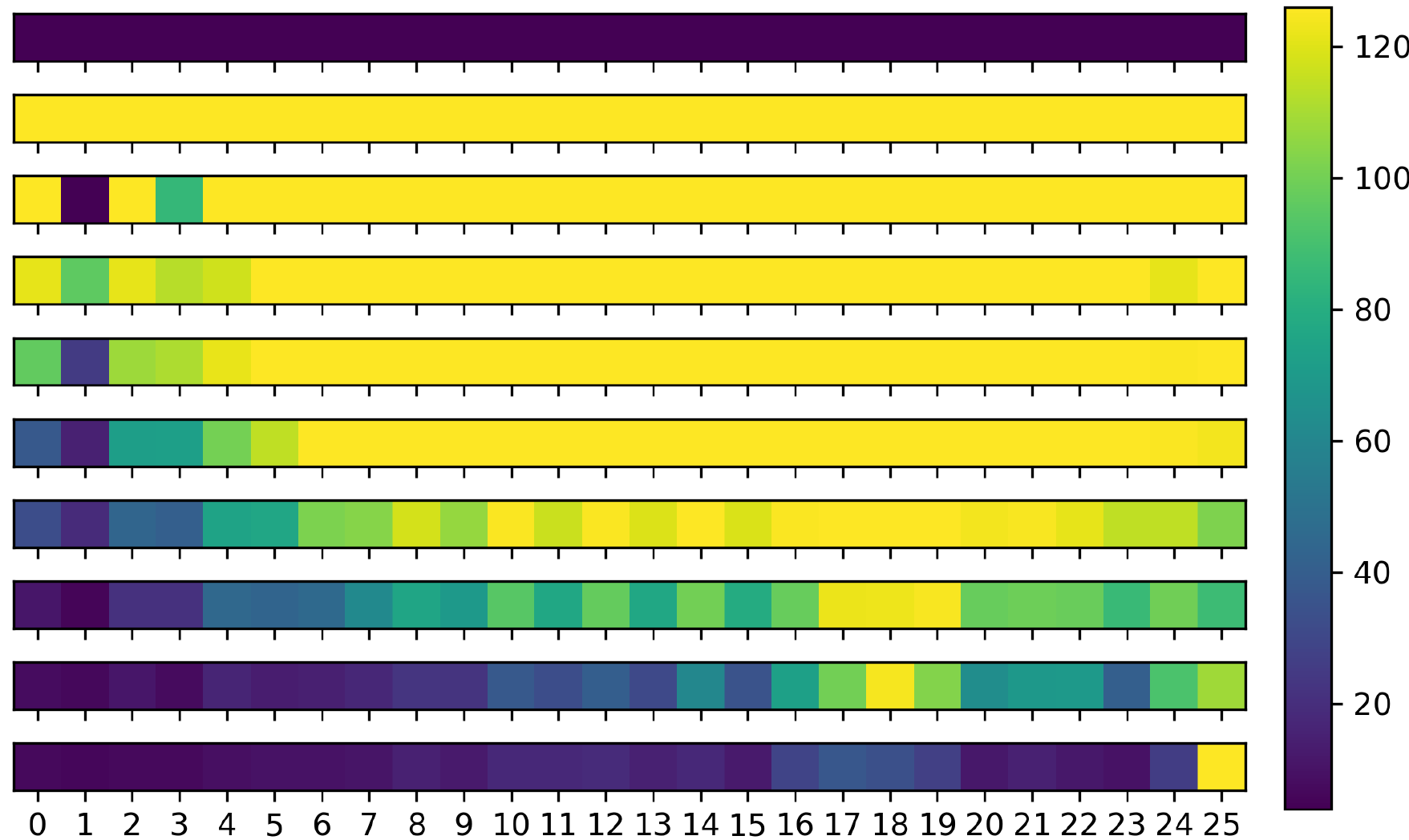}
        \caption[]%
        {{\small Math QA}}    
        \label{fig:mean and std of net34}
    \end{subfigure}
    \caption[]
    {\small The effective rank of weight update matrix. Brighter colors in the colorbar indicate a higher rank. The x-axis is the layer index, the y-axis is the training stage--with the earliest stages at the top and the latest at the bottom. Cognitively demanding tasks like MathQA, involve high-rank weight updates in almost all layers.} 
    \label{fig:rank}
\end{figure}

\subsection{Task Complexity Analysis}
\label{sec:exp_rank}
\paragraph{Setup.} Given our finding that more cognitively demanding tasks--i.e., tasks that require thinking steps-- require a larger number of layers to grow, we now investigate the underlying reason for this phenomenon. We analyze this from the perspective of the weight update matrix, hypothesizing that they induce higher-rank\footnote{high rank here is still a fraction of what a full rank update would be.} weight updates across a broader set of layers. To test this, we measure the change in the first MLP layer weights (up-projection) between vanilla model (SFT, without growing) checkpoints $\Delta \textbf{W}^{(1)} = \textbf{W}^{(1)}_t - \textbf{W}^{(1)}_{t-1}$ with $t$ as the training step; then calculate the effective rank of $\Delta \textbf{W}^{(1)}$.

\paragraph{Results.} Figure~\ref{fig:rank} supports our hypothesis by visualizing the rank of $\Delta \textbf{W}^{(1)}$ (lighter colors indicate higher rank) across all layers (x-axis) over time. For tasks like science entailment and translation, the high-rank updates are consistently \textbf{localized} to a specific subset of layers (e.g., early layers for entailment, middle-to-late layers for translation). In stark contrast, for MathQA, the high-rank updates are \textbf{distributed} broadly across nearly all layers, particularly in the initial fine-tuning stages. This suggests that simpler tasks can be learned by modifying a small, specialized part of the network, whereas complex reasoning tasks require diffuse, high-rank changes across the entire model, explaining why growing more layers is necessary for optimal performance.

\subsection{Preserving Base Model Function Vectors}
\label{sec:exp_fv}
\begin{table}[htp!]
\caption{Substantially Preserves Model Function Vector}
\label{tab:fv_similarity}
\small
\centering
\begin{tabular}{lccc}
    \toprule
    Dataset & Method & \# Intersecting Heads ($\uparrow$) & FV Similarity ($\uparrow$) \\
    \midrule
    \multirow{2}{*}{Entailment} & SFT & 2/10 & 0.28 \\
    & Ours & 5/10 & \textbf{0.95} \\
    \midrule
    \multirow{2}{*}{Translation} & SFT & 3/10 & 0.58 \\
    & Ours & 5/10 & \textbf{0.76} \\
    \bottomrule
    
\end{tabular}
\end{table}
\paragraph{Setup.} We further investigate \textit{how} our method mitigates forgetting by measuring its effect on the model's internal representations. This experiment is motivated by the finding that forgetting occurs when a fine-tuned model's latent representations diverge significantly from the original model's for the same set of inputs \citep{ramasesh2020anatomy, jiang2025unlocking}.
Following \citet{jiang2025unlocking}, we measure this impact using Function Vectors (FV): a compact vector
representation identified within transformer models hidden states during in-context learning (ICL) \cite{todd2024function}. First, an activation patching procedure \cite{meng2022locating} is performed to determine the causal set of attention heads important for the task. Next, FV is calculated by summing the activations from these key heads during ICL. More details on FV can be found in \cite{todd2024function}, and we provide a discussion section on FV in Appendix \ref{sec:appendix_fv}.

We then compare the original pre-trained model against models fine-tuned with both our method and standard fine-tuning (SFT). We assess the effects on FVs in two ways: (1) the number of causal heads overlap between the pre-trained and fine-tuned models, and (2) the cosine similarity between the pre-trained model's FV and the fine-tuned model's FV. For both metrics, higher values signify better preservation of the model's original capabilities.



\paragraph{Results.}
Table \ref{tab:fv_similarity} shows that our method successfully preserves the model's internal representation, unlike standard fine-tuning (SFT). Our approach achieved a high FV cosine similarity of 0.95 and retained 5 causal attention heads. In contrast, SFT's similarity dropped to 0.28, with only 2-3 heads overlapping. These results indicate our method effectively prevents the representational drift associated with catastrophic forgetting.

\section{Related Work}
\label{sec:related_work}

\paragraph{Catastrophic Forgetting.}
Catastrophic forgetting is a long-standing challenge in machine learning \citep{mccloskey1989catastrophic, french1999catastrophic}. One popular approach is using regularization to constrain model updates, adding a penalty to prevent the fine-tuned parameters from deviating significantly from their original state \citep{kirkpatrick2017overcoming, lee2017overcoming, goodfellow2013empirical, li2019learn, pmlr-v80-serra18a, NEURIPS2018_f31b2046}. A second popular approach is replay, where a subset of past data is stored and revisited during finetuning to reinforce old knowledge \citep{rolnick2019experience, ratcliff1990connectionist, rebuffi2017icarl, lopez2017gradient, jung2016less, li2017learning}. Both approaches operate under specific, constrained settings \citep{kemker2018measuring} and force an inherent trade-off within a fixed-capacity model. This work presents an alternative path that circumvents this trade-off by expanding model capacity.

\paragraph{Network Growing}
Network growing methods add capacity to pre-trained models, with early work initializing larger networks from smaller ones \citep{rusu2016progressive, zhang2020class, wei2016network, shen2022staged, mazzawi2023deep}. Recent approaches for Transformer architectures prioritize function-preserving expansions to ensure stable fine-tuning \citep{gesmundo2023composable}, but this has created a dilemma. Methods that achieve stability add zero-initialized weights, forcing new skills to be learned from scratch \citep{han2025loire, wu2024llama}. While approaches that reuse knowledge by copying entire blocks typically violate the function-preserving constraint \citep{wei2025control}. Thus, a method that can both leverage pre-trained knowledge and guarantee function preservation remains a critical gap in the literature. We address this gap with a novel expansion method, inspired by \citet{mazzawi2023deep}, that is designed to be both function-preserving and exploiting pre-trained weights.

\paragraph{Parameter Efficient Finetuning (PEFT).} PEFT techniques aim to adapt models to new tasks by updating only a small fraction of parameters. Prominent methods include Adapters \citep{houlsby2019parameter}, which insert small, task-specific modules into a frozen model, and Low-Rank Adaptation (LoRA) \citep{hu2022lora}, which approximates weight updates using trainable low-rank matrices. The success of LoRA has inspired a large family of variants that improve its performance or extend it to specific settings like model quantization \citep{hayou2024lora+, zhang2023lora, valipour2022dylora, liu2024dora, zi2023delta, kopiczko2023vera, zhang2023adalora, dettmers2023qlora, zhou2024lora}. Our work is \textit{orthogonal} to this line of research. While PEFT methods focus on efficiency, our primary goal is to eliminate catastrophic forgetting while matching the performance of full fine-tuning. Our approach can be readily combined with PEFT techniques to gain both efficiency and knowledge preservation.

\section{Conclusion}
\label{sec:conclusion}

We present a method that resolves the fundamental trade-off between learning new tasks and retaining old knowledge in pre-trained models. Our approach adds capacity to the model's MLP layers in a function-preserving manner, ensuring training stability. Our experiments prove that this approach completely eliminates catastrophic forgetting, matching standard fine-tuning performance on new tasks while fully preserving original capabilities. Our approach is modular and can be applied only to a subset of layers and still match full fine-tuning performance, enabling computational saving. Finally, we show that performance scales with the amount of added capacity.

\bibliographystyle{plainnat}
\bibliography{main}
\newpage
\appendix
\section{Glossary} 
Table \ref{table:glossary} shows glossary of terms used in this paper.
\label{appendix:glossary}
\label{sec:gloss}
\begin{table*}[htp!]
\centering
\begin{tabular}{l l}
\toprule
Symbol & Definition \\
\midrule
$M_0$ & pretrained model \\
$\theta_0$ & pretrained model parameters \\
$M_T$ & finetuned model \\
$\theta_T$ & finetuned model parameters \\
$\mathcal{L}_T$ & finetuning task loss \\
$\mathcal{D}_{T}$ & downstream task dataset  \\
$\mathcal{D}_{\text{proxy}}$ & proxy dataset to measure performance on pretraining set \\
$N$ & Number of model layers \\
$n$ & model layer index \\
$\textbf{X}$ & input to model MLP module \\
$s$ & input sequence length \\
$h$ & model hidden dimension \\
$p$ & MLP hidden dimension \\
$\textbf{W}_n^{(1)}$ & weight matrix of first layer of MLP module (up-projection) \\
$\textbf{W}_n^{(2)}$ & weight matrix of second layer of MLP module (down-projection) \\
$\textbf{B}_n^{(1)}$ & bias term of first layer of MLP module (up-projection) \\
$\textbf{B}_n^{(2)}$ & bias term of second layer of MLP module (down-projection) \\
$\hat{\textbf{W}}_n^{(1)}$ & expanded matrix of first layer of MLP module (up-projection) \\
$\hat{\textbf{W}}_n^{(2)}$ & expanded matrix of second layer of MLP module (down-projection) \\
$\textbf{Y}$ & outpaut of MLP module \\
\bottomrule
\end{tabular}
\caption{
	Glossary of variables and symbols used in this paper.
}
\label{table:glossary}
\end{table*}

\section{Function vectors}
\label{sec:appendix_fv}

This section provides a brief overview of \textbf{Function Vectors (FVs)} \citep{todd2024function}, a method for extracting a compact vector representation of a task from a Transformer's hidden states. FVs are a powerful tool for analyzing how models perform in-context learning (ICL) and have become valuable for understanding catastrophic forgetting. A significant change to a task's FV after fine-tuning can signal that the model's underlying neural circuit for that task has been damaged or erased \citep{jiang2025unlocking, ramasesh2020anatomy}.

The core idea is that a small subset of attention heads---the \textbf{causal heads}---can be identified as being most critical for performing a specific task. The FV is then constructed by summing the average activations of only these causal heads, which effectively isolates the functional circuit for that task.

The derivation of a function vector $\theta_t$ for a given task $t$ and model $f$ involves three main steps:

\paragraph{Step 1: Compute Mean Clean Activations.}
The process begins by running the model on a dataset $P_t$ of clean, task-specific in-context learning prompts. For each attention head $a_{lj}$ (at layer $l$, head $j$), an average activation vector is computed across all these clean prompts. This yields the head's activation pattern when performing the task correctly, denoted as $\bar{a}_{lj}^t$.
\[
\bar{a}_{lj}^t = \frac{1}{|P_t|} \sum_{p_i^t \in P_t} a_{lj}(p_i^t)
\]

\paragraph{Step 2: Identify Causal Heads via Activation Patching.}
Next, \textbf{activation patching} (also known as causal tracing) \citep{meng2022locating} is used to identify the heads most responsible for the task. The procedure is as follows:
\begin{enumerate}
    \item \textbf{Corrupted prompts} $\tilde{p}_i^t$ are created by shuffling the input-output pairs, causing the model to fail at the task.
    \item The model is then run on these corrupted prompts. During the forward pass, an intervention is performed on a single attention head $a_{lj}$ at a time by replacing its corrupted activation with its clean mean activation $\bar{a}_{lj}^t$ from Step 1.
    \item Finally, the \textbf{Causal Indirect Effect (CIE)} is measured. This value quantifies how much the patch restores the model's ability to produce the correct answer $y_i$. A high CIE indicates the head is causally important.
\end{enumerate}
\[
\text{CIE}(a_{lj}) = \mathbb{E}_{p_i^t \in P_t} \left[ f(\tilde{p}_i^t | a_{lj} \leftarrow \bar{a}^t_{lj})[y_i] - f(\tilde{p}_i^t)[y_i] \right]
\]
The set of causal heads, $\mathcal{S}$, is defined as the top-k heads (e.g., top-10) with the highest CIE scores.

\paragraph{Step 3: Construct the Function Vector.}
The function vector $\theta_t$ is then constructed by summing the mean clean activations of the causal heads identified in the previous step.
\[
\theta_t = \sum_{(l,j) \in \mathcal{S}}\bar{a}_{lj}^t
\]

In Section \ref{sec:exp_fv} of our main paper, we use FVs to demonstrate that our approach preserves latent representation of the pretrained model. We show that after fine-tuning, our method \textbf{retains 5 of the top 10 original causal heads} for a task and maintains a \textbf{function vector cosine similarity of 0.95} with the original model. This provides strong evidence that our method mitigates catastrophic forgetting by preserving the essential computational circuits of the base model. For a more detailed treatment of FVs, we refer readers to \citet{todd2024function} and \citet{jiang2025unlocking}.

\section{Supplementary Experiments}
\label{sec:appendix_extra_exp}
This section provides additional experiments to further validate our claims. Specifically:
\begin{itemize}
    \item \textbf{Our results persist in  larger models (\ref{sec:exp_scale}).}
    \item \textbf{Our proposed growing strategy is superior to a zero-initialization (\ref{sec:exp_zero}).}
    \item \textbf{Growing MLP is more effective than growing attention (\ref{sec:exp_attn}).}
\end{itemize}

\subsection{Validation on Larger-Scale Models}
\label{sec:exp_scale}
\begin{figure}[htp!]
        \centering
        \begin{subfigure}[b]{0.49\textwidth}
            \centering
            \includegraphics[width=\textwidth]{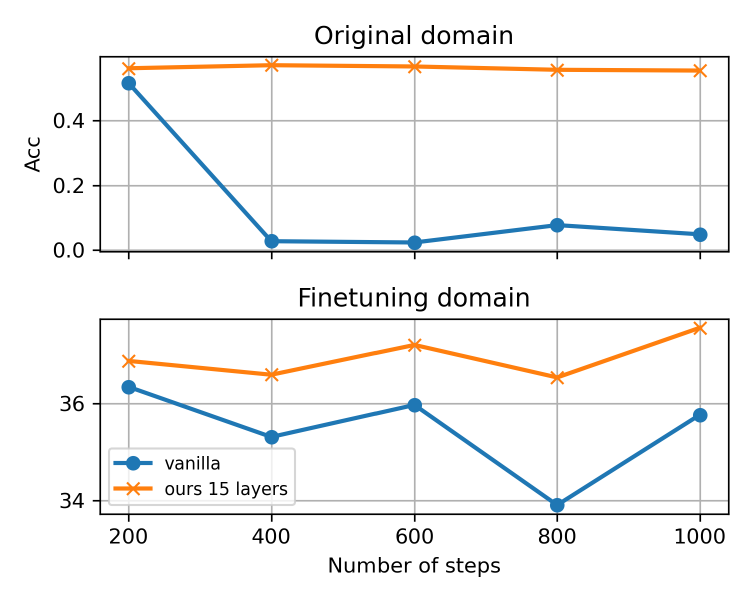}
            \caption[]%
            {{\small French translation}}    
            \label{fig:4b_mtnt}
        \end{subfigure}
        \begin{subfigure}[b]{0.49\textwidth}  
            \centering 
            \includegraphics[width=\textwidth]{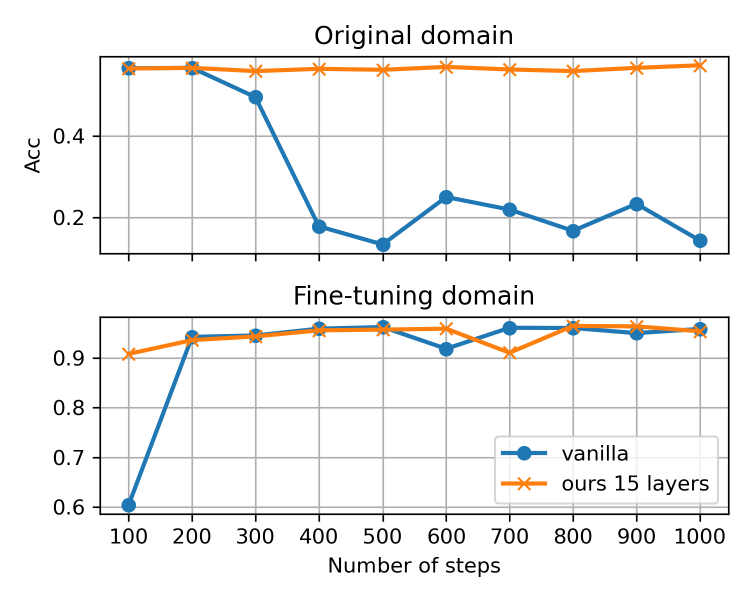}
            \caption[]%
            {{\small Science entailment}}    
            \label{fig:4b_scitail}
        \end{subfigure}
        \begin{subfigure}[b]{0.49\textwidth}   
            \centering 
            \includegraphics[width=\textwidth]{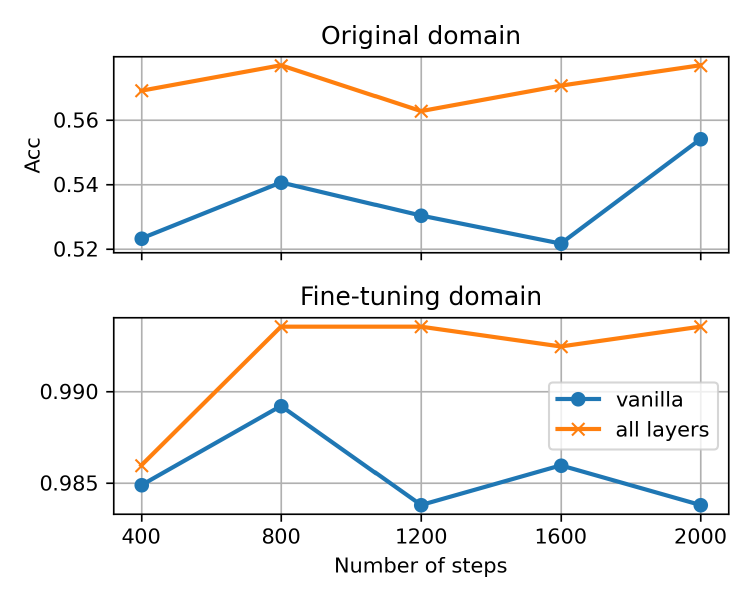}
            \caption[]%
            {{\small Science Q\&A}}    
            \label{fig:4b_qasc}
        \end{subfigure}
        \begin{subfigure}[b]{0.49\textwidth}   
            \centering 
            \includegraphics[width=\textwidth]{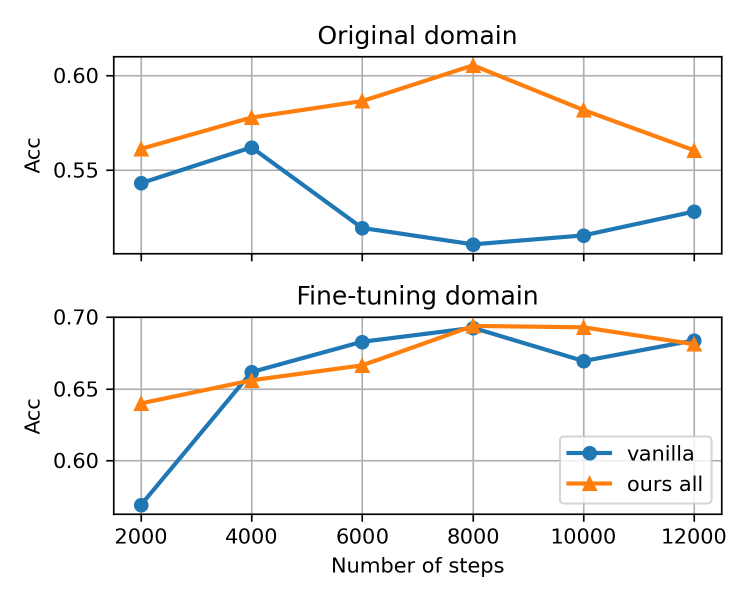}
            \caption[]%
            {{\small Math QA}}    
            \label{fig:4b_math}
        \end{subfigure}
        \caption[]
        {\small Our results hold on larger models.}
        \label{fig:result_4b}
    \end{figure}
To verify that our results generalize to larger models, we run our experiments on Gemma-4B models using the setup from Section \ref{sec:experiments}. As shown in Figure \ref{fig:result_4b}, our method scales effectively: on larger models (4x the parameters of the model used in Section \ref{sec:experiments}, it completely preserves original task performance while matching the new task performance of standard fine-tuning.

\subsection{Ablation Study on Initialization Strategy}
\label{sec:exp_zero}

\begin{figure}[htp!]
    \begin{subfigure}{0.49\textwidth}
        \centering
        \includegraphics[width=\textwidth]{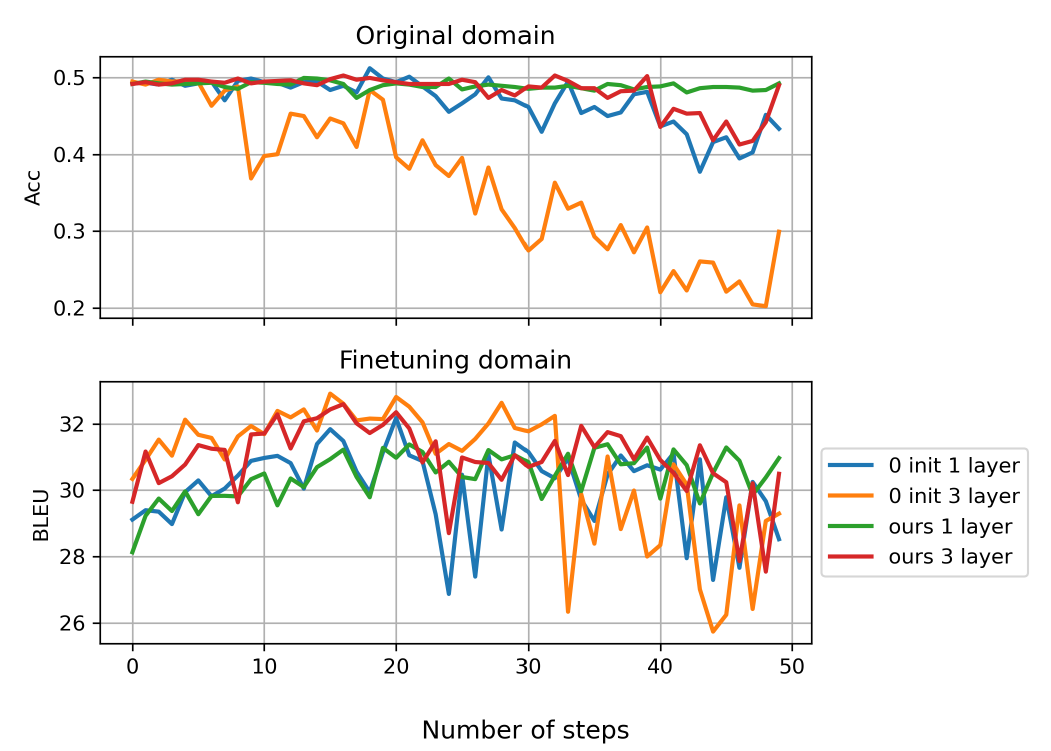}
        \caption[]%
        {{\small Zero initialization, train old and new parameters}}    
        \label{fig:zero_train}
    \end{subfigure}
    \begin{subfigure}{0.49\textwidth}  
        \centering 
        \includegraphics[width=\textwidth]{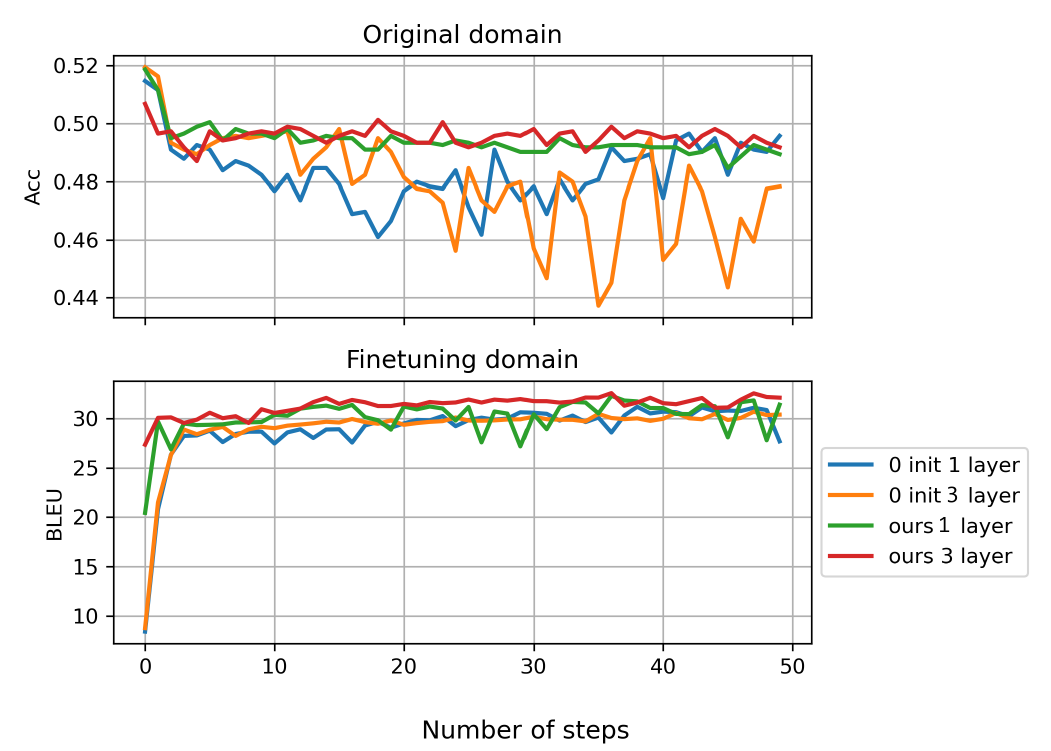}
        \caption[]%
        {{\small Zero initialization, train only new parameters}}    
        \label{fig:zero_freeze}
    \end{subfigure}
    \caption[]
    {\small Comparison with zero initialization} 
    \label{fig:zero_comparison}
\end{figure}

To isolate the benefits of our proposed growth strategy, we compare it against a naive but common baseline: adding new layers with zero-initialization. We evaluate this baseline in two distinct settings to test its impact on both knowledge retention and new task acquisition.

First, when all parameters (old and new) are trained jointly, the zero-initialized layers fail to effectively learn the new task. Figure \ref{fig:zero_train} (bottom) shows that this baseline significantly underperforms our method on the fine-tuning domain. Second, in a setting analogous to our own, where original parameters are frozen, the zero-initialized model suffers from severe catastrophic forgetting. Figure \ref{fig:zero_freeze} (top) shows that performance on the original task degrades sharply as more layers are added.

These results demonstrate that simply adding new capacity is insufficient. Our initialization strategy is critical for both preserving prior knowledge and enabling the new parameters to learn the target task efficiently.


\subsection{Ablation on Component to Grow}
\label{sec:exp_attn}

\begin{figure}[htp!]
    \begin{subfigure}{0.49\textwidth}
        \centering
        \includegraphics[width=\textwidth]{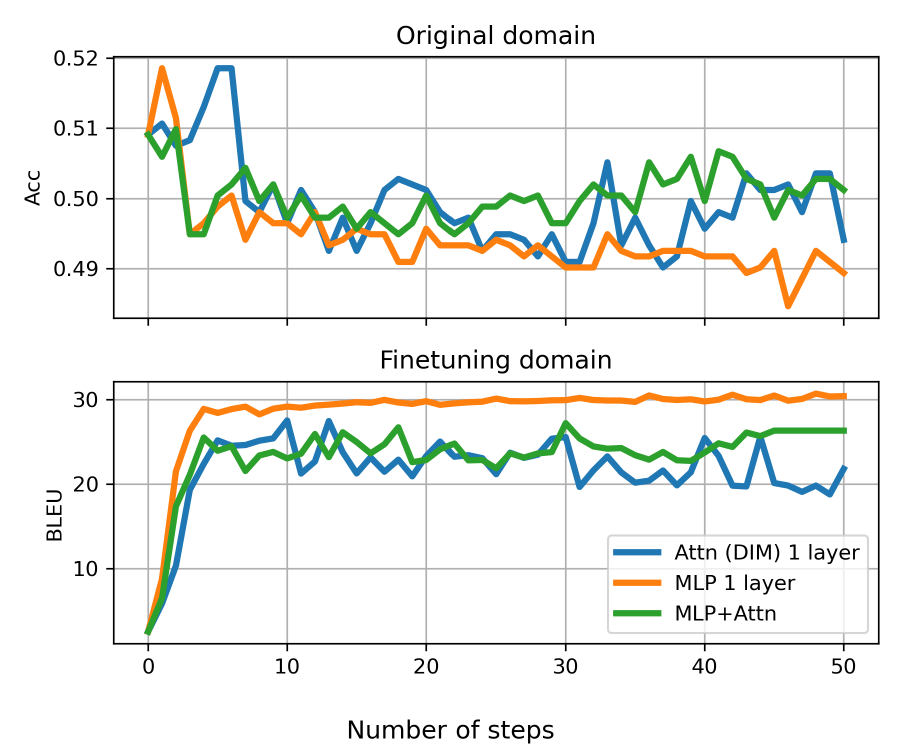}
        \caption[]%
        {{\small Grow attention head dimension}}    
        \label{fig:attn_dim}
    \end{subfigure}
    \begin{subfigure}{0.49\textwidth}  
        \centering 
        \includegraphics[width=\textwidth]{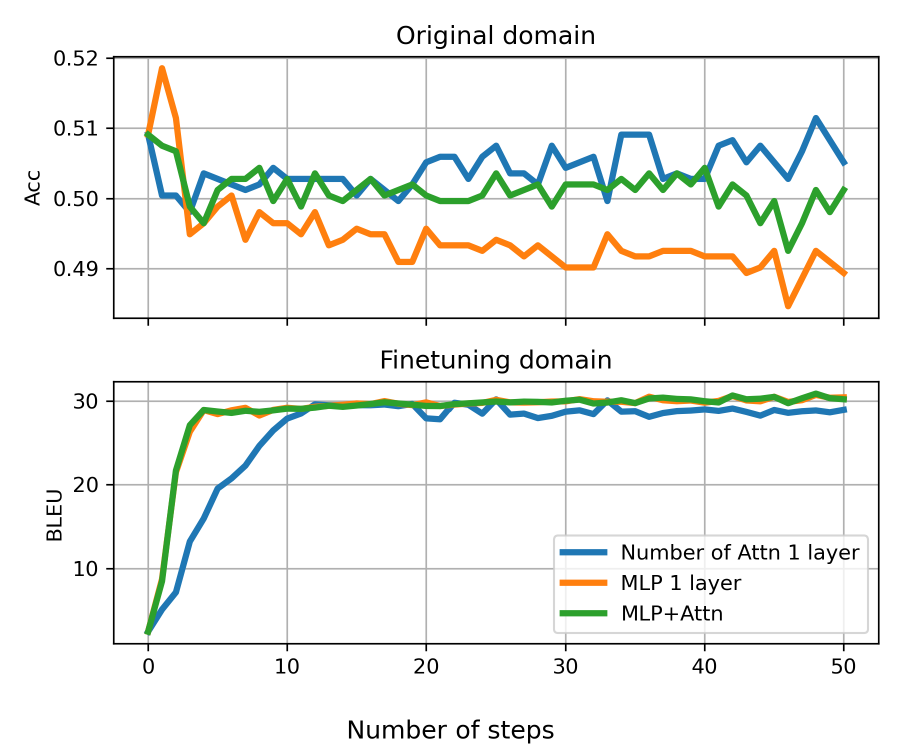}
        \caption[]%
        {{\small Grow number of attention heads}}    
        \label{fig:n_attn}
    \end{subfigure}
    \caption[]
    {\small Comparison with growing attention head module} 
    \label{fig:attn_comparison}
\end{figure}

To justify our focus on growing MLP layers, we conducted an ablation study to determine which part of the Transformer block is most effective to grow. We compare our approach against two alternatives for growing the attention mechanism: (1) increasing the attention head dimension (Figure \ref{fig:attn_dim}) and (2) increasing the number of attention heads (Figure \ref{fig:n_attn}).

Our findings clearly show that both strategies for expanding the attention module yield inferior performance compared to growing the MLP module. More surprisingly, we found that a combined approach of growing both the MLP and attention components simultaneously offered no additional benefit over growing MLP alone. This result provides empirical validation for our design choice to target the MLP layers for expansion, as this is the most parameter-efficient and performant strategy.

\end{document}